%% file: acl_latex.tex
\pdfoutput=1

\documentclass[11pt]{article}

\usepackage[]{acl}

\usepackage{times}
\usepackage{latexsym}

\usepackage[T1]{fontenc}

\usepackage[utf8]{inputenc}

\usepackage{microtype}

\usepackage[linesnumbered, ruled]{algorithm2e}
\usepackage[T1]{fontenc}
\usepackage{dsfont}

\usepackage{mdframed}
\usepackage{booktabs}
\usepackage{graphicx}

\usepackage[normalem]{ulem}
\usepackage{enumitem}
\usepackage{xcolor}
\usepackage{soul}
\colorlet{soulred}{red!30}
\sethlcolor{soulred}%

\usepackage{amsmath,amsthm,amsfonts,amssymb,bm}
\usepackage{framed}
\usepackage{varioref}
\usepackage{float}
\usepackage{multirow}
\usepackage{dashrule}
\usepackage{url}
\usepackage{pifont}
\usepackage{lipsum}
\usepackage{algorithmic}

\usepackage{multicol}

\title{Unsupervised Slot Schema Induction for Task-oriented Dialog}

\author{Dian Yu\textsuperscript{1}\thanks{work done during internship at Google Research}, Mingqiu Wang\textsuperscript{2}, Yuan Cao\textsuperscript{2}, Izhak Shafran\textsuperscript{2}, Laurent El Shafey\textsuperscript{2}, Hagen Soltau\textsuperscript{2}\\
\textsuperscript{1} University of California, Davis\\
\textsuperscript{2} Google Research\\
\texttt{dianyu@ucdavis.edu}\\
\texttt{\{mingqiuwang, yuancao, izhak, shafey, soltau\}@google.com}
}

\begin{document}
\maketitle
\begin{abstract}
Carefully-designed schemas describing how to collect and annotate dialog corpora are a prerequisite towards building task-oriented dialog systems. 
In practical applications, manually designing schemas can be error-prone, laborious, iterative, and slow, especially when the schema is complicated. To alleviate this expensive and time consuming process, we propose an unsupervised approach for slot schema induction from unlabeled dialog corpora. 
Leveraging in-domain language models and unsupervised parsing structures, our data-driven approach extracts candidate slots without constraints, followed by coarse-to-fine clustering to induce slot types.
We compare our method against several strong supervised baselines, and show significant performance improvement in slot schema induction on MultiWoz and SGD datasets. We also demonstrate the effectiveness of induced schemas on downstream applications including dialog state tracking and response generation.

\end{abstract}

\section{Introduction}
Defining task-specific schemas, including intents and arguments, is the first step of building a task-oriented dialog (TOD) system.
In real-world applications such as call centers, we may have abundant conversation logs from real users and system assistants without annotation. 
To build an effective system, experts need to study thousands of conversations, find relevant phrases, manually group phrases into concepts, and iteratively build the schema to cover use cases. The schema is then used to annotate belief states and train models.
This process is labor-intensive, error-prone, expensive, and slow~\cite{eric-etal-2020-multiwoz, zang-etal-2020-multiwoz, min-etal-2020-dialogue, yu-yu-2021-midas}. 
As a prerequisite, it hinders quick deployment for new domains and tasks. We therefore are interested in developing automatic schema induction methods in this work to create the ontology\footnote{We use ``schema'' and ``ontology'' interchangeably in this paper. Following previous work in literature, we focus on schema induction for slots, which is more challenging than domains and intents.} from conversations for TOD tasks.

\begin{figure}[t]
\centering
\includegraphics[width=\linewidth]{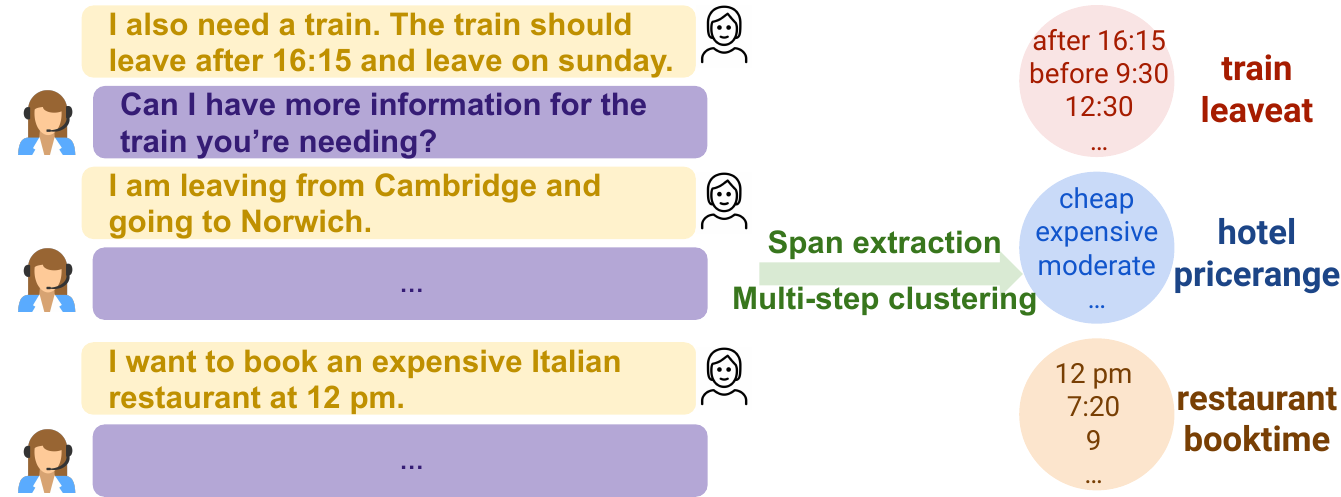}
\vspace{-1em}
\caption{Overview of slot schema induction from raw conversations. We use a bottom-up representation level distance function derived from
pre-trained LMs (combined with PCFG structure) to extract informative candidate phrases such as ``after 16:15'' and ``expensive''. The spans are subsequently clustered through multiple stages to form coarse to fine categories.
The ground truth mapping is shown on the right (such as ``train leaveat'').}
\label{fig:illustration}
\vspace{-1em}
\end{figure}

Most existing approaches for slot schema induction rely on syntactic or semantic models trained with labeled data \cite{chen-etal-2013-unsupervised, hudecek-etal-2021-discovering, min-etal-2020-dialogue}. Our proposed method, on the other hand, is completely unsupervised without requiring generic parsers and heuristics, and hence portable to new tasks and domains seamlessly, overcoming the limitations of previous research. 
Analogous to human experts, 
our procedure is divided into two general steps: relevant span extraction, and slot categorization. Fig.~\ref{fig:illustration} provides an overview of our approach.
We introduce a bottom-up span extraction method leveraging 
a pre-trained language model (LM) and regularized by unsupervised probabilistic context-free grammar (PCFG) structure.
We also propose a multi-step auto-tuned clustering method to group the extracted spans into fine-grained slot types with hierarchy. 

We demonstrate that our unsupervised induced slot schema is well-aligned with expert-designed reference schema on the public MultiWoZ \cite{budzianowski-etal-2018-multiwoz} and SGD \cite{rastogi-etal-2020-towards} datasets.
We further evaluate the induced schema on dialog state tracking (DST) and response generation to indicate usefulness and demonstrate performance gains over strong supervised baselines. Meanwhile, our method is applicable to more realistic scenarios with complicated schemas.

%

\section{Related Work} \label{sec:background}
Schema induction from dialog logs has not been studied extensively in the literature and developers resort to a patch work of tools to automate parts of the process. We first introduce related work on schema induction for dialog, and then discuss previous research on span extraction as part of schema induction for slots.

\paragraph{Schema induction for dialog} 
Motivated by the practical advantages of unsupervised schema induction
such as reducing annotation cost and avoiding human bias,
\citet{klasinas-etal-2014-semeval, athanasopoulou-etal-2014-using} propose to induce spoken dialog grammar based on n-grams
to generate fragments.
Different from studying semantic grammars, \citet{chen-etal-2013-unsupervised, chen-etal-2014-frame,  chen-etal-2015-semantic, chen-etal-2015-jointly, hudecek-etal-2021-discovering} propose to utilize annotated FrameNet~\cite{baker-etal-1998-berkeley} to label semantic frames for raw utterances \cite{das-etal-2010-probabilistic}. The frames are designed on generic semantic context, which contains frames that are related to the target domain (such as "expensiveness") and irrelevant (such as "capability"), while other relevant slots such as ``internet'' cannot be extracted because they do not have corresponding defined frames. This line of work focuses on ranking extracted frame clusters and then manually maps the top-ranked induced slots to reference slots. Instead of FrameNet, \citet{shi-etal-2018-auto} extract features such as noun phrases (NPs) using part-of-speech (POS) tags and frequent words and aggregate them via a hierarchical clustering method, but only about 70\% target slots can be induced. 
In addition to the unsatisfactory induction results due to candidate slot extraction, most of the previous works are only applicable to a single domain such as restaurant booking
with a small amount of data, and require manual tuning to find spans and generate results.
These methods are not easily adaptable to unseen tasks and services.

The most comparable work to ours is probably \citet{min-etal-2020-dialogue}, which is not bounded by an existing set of candidate values so that potentially all slots can be captured. They propose to mix POS tags, named entities, and coreferences with a set of rules to find slot candidates while filtering irrelevant spans using manually updated filtering lists. In comparison, our method does not require any supervised tool and can be easily adapted to new domains and tasks with self-supervised learning. In addition to flexibility, despite our simple and more stable clustering process compared to their variational embedding generative approach \cite{jiang-etal-2017-variational}, our method achieves better performance on slot schema induction and our induced schema is more useful for downstream tasks. 

We survey schema induction work for other natural language processing tasks in Appendix \ref{app:related_work}.

\paragraph{Span extraction} 

Previous works in span extraction consider all combination of tokens 
as candidates \cite{yu-etal-2021-shot}.
Alternatively, keyphrase extraction research \cite{campos-etal-2018-yake, bennani-smires-etal-2018-simple} mostly depends on corpus statistics (such as frequency), similarity between phrase and document embeddings, or POS tags \cite{wan-xiao-2008-single, liu-etal-2009-clustering}, and formulates the task as a ranking problem. Although these methods can find meaningful phrases, they may result in a low recall for TOD settings. For instance, 
the contextual semantics of a span (such as time) in an utterance may not represent the utterance-level semantics compared to other generic phrases. Other methods for span extraction include syntactic chunking, but mostly require supervised data \cite{li-etal-2021-neural} and heuristics (such as considering ``noun phrases'' or ``verb phrases''), and thus are not flexible and robust compared to our method. 

Finally, target spans can be found in syntactic structures which can be potentially induced from supervised parsers or unsupervised grammar induction \cite{klein-manning-2002-generative, klein-manning-2004-corpus, shen-etal-2018-neural, drozdov-etal-2019-unsupervised, zhang-etal-2021-video}. 
\citet{kim-etal-2020-are} probe LMs and observe that recursively splitting sentences into binary trees in a top-down approach can correlate to constituency parsing. However, 
unlike the task of predicting relationship between words in a sentence where phrases at each level of a hierarchical structure are valid, detecting clear boundaries is critical to span extraction but challenging with various phrase lengths. 
Even though more flexible compared to semantic parsers that are limited by pre-defined roles, there is no straightforward way to apply these methods to span extraction.

\section{Unsupervised Slot Schema Induction}
\label{sec:method}
Our proposed method for slot schema induction consists of a fully unsupervised span extraction stage followed by coarse-to-fine clustering. 
The resulting clusters can be mapped to slot type labels.

\subsection{Overview}
Given user utterances
from raw conversations, our goal is to induce the schema of slot types $\mathcal{S}$ and their corresponding slot values.
The span extraction stage extracts spans (e.g., ``with wifi'') from an utterance $\mathbf{x}$.
The candidate spans from all user utterances are then clustered into a set of groups $\mathcal{S}$ where each group $s_i$ corresponds to a slot type such as ``internet'' with values ``with wifi'', ``no wifi'', and ``doesn't matter''.
The induced slot schema can be later used for downstream applications such as dialog state tracking and response generation.

\input{span_extract_alg}

\subsection{Candidate span extraction}
\label{sec:span}


\paragraph{Challenges} 
Since it is unclear what spans are meaningful phrases representative of task-specific slots, candidate span extraction presents two challenges. Firstly, with either supervised or unsupervised predicted structures, there is no protocol on what constituent and from what level we should extract the spans from without relying on dataset-specific heuristics, especially as structured representations are often compositional \cite{herzig-berant-2021-span}. 
The second challenge is that span extraction methods should be flexible and robust to unseen tasks and domains. To tackle these problems, we leverage pre-trained LMs and propose a novel bottom-up attention-based span extraction method regularized by unsupervised PCFG for better structure representation. Because our method does not need any supervised data, the second problem can be effectively addressed by in-domain self-training.
The full algorithm is outlined in Algorithm~\ref{alg:span_extract}.

\noindent



\begin{figure}[t]
\centering
\includegraphics[width=\columnwidth]{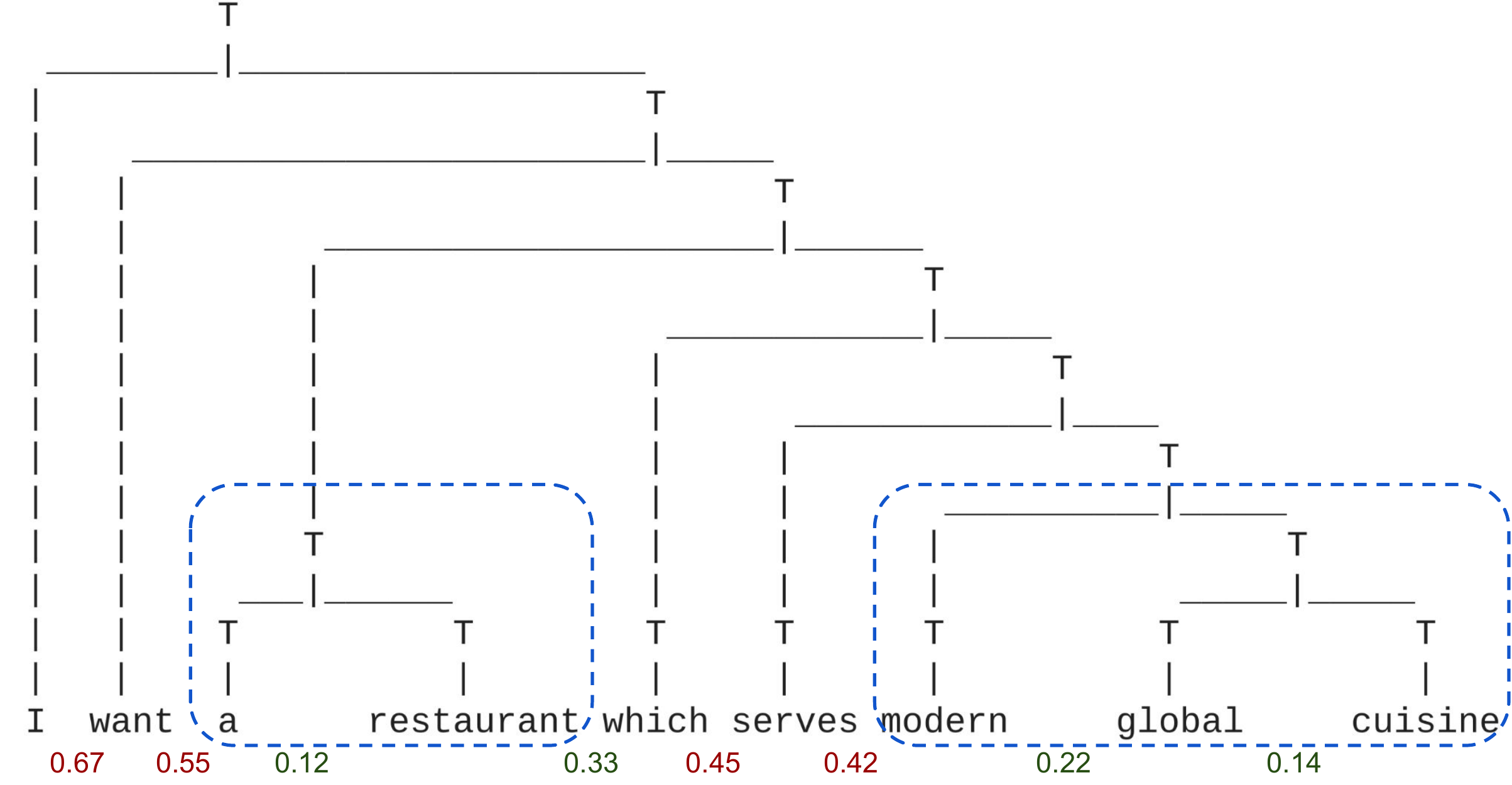}
\vspace{-1em}
\caption{
Illustration of span extraction where LM-derived distance function (distances between tokens are shown below the text) is constrained by a structure predicted by PCFG (tree structure shown in the figure).
Numbers in red are above the median threshold ($0.375$) while numbers in green are below, indicating that the tokens share similar semantics and are from the same span. We can then extract candidate phrases ``a restaurant'' and ``modern global cuisine'', together with unigrams ``I'', ``want'', ``which'', and ``serves''.}
\label{fig:pcfg_ex}
\vspace{-1em}
\end{figure}

\noindent \paragraph{Bottom-up attention-based extraction with LMs and PCFG regularization} 
Recent studies reveal that attention distributions in pre-trained LMs
can indicate syntactic relationships among tokens \cite{clark-etal-2019-bert}.
Therefore, we hypothesize that
similar attention distributions indicate tokens to form a meaningful phrase.
We define the distance between attention distributions as a symmetric Jensen-Shannon divergence \cite{clark-etal-2019-bert}, and iteratively merge tokens whose distance is smaller than a threshold\footnote{We use the median of all pairwise distances in an utterance in the experiments. We also compared other thresholds such as mean but did not observe significant difference.} in a bottom-up fashion. We start from the smallest distance to the largest, where the merged tokens are considered as a new token in the next iteration but the distribution distance with adjacent tokens remains the same. Fig.~\ref{fig:pcfg_ex} illustrates the distances between tokens from a pre-trained LM for an example sentence where adjacent tokens such as ``global'' and ``cuisine'' are merged but not ``serves'' and ``modern''. This new decoding method enables us to effectively group tokens into phrases with precise boundaries.

Although LMs can be used to induce grammar, their training objectives are not optimized for sentence structure prediction, hence falling behind unsupervised PCFG \cite{kim-etal-2020-are} on syntactic modeling\footnote{Note that although \citet{kim-etal-2020-are} utilizes attention to construct constituency trees, our methods are different: we propose a bottom-up approach to identify meaningful spans with clear boundaries (rather than a tree structure) after in-domain further pre-training.}. Utilizing attention distribution from LM representations to extract spans can thus be fuzzy and noisy. We therefore employ unsupervised PCFG proposed by \citet{kim-etal-2019-compound} as a mechanism to regularize our bottom-up span extraction. 
Instead of relying solely on attention distribution, we in addition require two tokens to share the same parent in the predicted PCFG tree structure before merging.
This extra requirement reduces the noise from the distribution divergence in a sub-optimal structure representation. 
An example illustrating the necessity of span constraint is given in Fig.~\ref{fig:pcfg_ex}. Even though the distance between ``restaurant'' and ``which'' is small ($0.33$), we disregard this span since they do not belong to the same parent in the PCFG structure. After merging two tokens, we assign the grandparent of the two tokens as the new parent, and continue the iteration until all distances are examined.

\noindent \paragraph{Self-supervised in-domain training}
Our attention-based approach enables us to extract phrases beyond certain n-grams,
or certain types of phrases in a specific hierarchical layer. 
More importantly, 
it is appealing to adapt to new domains and services, where a LM can be further trained to encode structure representations without any annotated data and to group tokens into candidate phrases based on the training corpus.
To encourage efficient span extraction above token-level representation, we further pre-train a SpanBERT model \cite{joshi-etal-2020-spanbert} by predicting masked spans together with a span boundary objective (denoted as TOD-Span) on TOD data \cite{wu-etal-2020-tod}. In addition to masking random contiguous spans with a geometric distribution, we also mask spans inspired by recent findings such as segmented PMI \cite{levine-etal-2021-pmimasking}
among other methods (See Appendix~\ref{app:tod_dst} for details). This process can be thought of as incorporating corpus statistics such as phrase frequency into the model implicitly \cite{henderson-vulic-2021-convex}.

The unsupervised PCFG is trained to maximize the marginal likelihood of in-domain utterances with the inside-outside algorithm on the same TOD dataset. 
Similar to self-supervised LMs, this process is flexible and robust against domain mismatch, a common problem with supervised parsers \cite{davidson-etal-2019-dependency}.
At inference time, the trained model predicts a Chomsky normal form from Viterbi decoding \cite{forney-1973-viterbi}.

\subsection{Clustering candidate spans}
\label{sec:cluster}
\paragraph{Challenges}
After extracting candidate spans as potential slot values, we apply contextualized clustering on them to form latent concepts each slot value belongs to.
We face two major challenges. Firstly, for any clustering method, hyperparameters such as the number of clusters are critical to the clustering quality, while they are not known for a new domain. Secondly, because of the trivial differences in slot types (for example, a location can be a ``train departure place'', or a ``taxi arrival place''), clustering
requires considering different dimensions of semantics and pragmatics. 
To address these problems, we propose an auto-tuned, coarse-to-fine multi-step clustering method. The pseudo code of the clustering algorithm can be found in Appendix~\ref{app:algorithm}.

\paragraph{Auto-tuning hyperparameters}
To avoid hyperparameter tuning, we utilize density-based HDBSCAN \cite{mcinnes-etal-2017-hdbscan}. 
Compared to other clustering methods such as K-Means, HDBSCAN is mainly parametrized by the minimum number of samples per cluster, and resulting clusters are known to be less sensitive to this parameter.
We set this number automatically by maximizing the averaged Silhouette coefficient \cite{rousseeuw-1987-silhouettes}
\begin{equation}
    s = \frac{b - a}{max(a, b)} \notag
\end{equation}
across all clusters where $a$ represents the distance between samples in a cluster, and $b$ measures the distance between samples across clusters.




\begin{figure}[t]
\centering
\includegraphics[width=\linewidth]{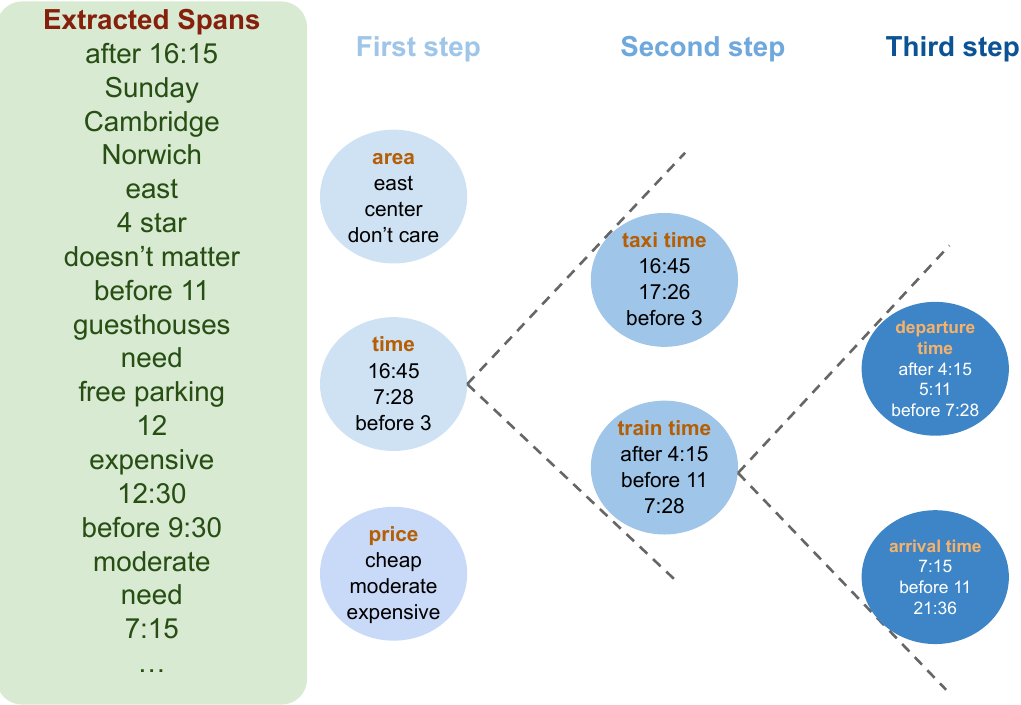}
\vspace{-1em}
\caption{Multi-step clustering procedure. Each coarse cluster is further refined by next-step clustering. The first step uses contextualized span representations to capture salient groups (such as a cluster about time), and the second step uses the utterance-level representations of each span to capture domain and intent information (such as the train service and taxi service). The third step utilizes span-level representation for fine-grained slot types. 
}
\label{fig:multistep_illustration}
\vspace{-1em}
\end{figure}

\paragraph{Multi-step clustering} 
The input to our first-step clustering is the contextualized span-level representation from the extracted spans. Specifically, we consider the mean representation of tokens in the span from the last layer as the span representation.
To prevent the surface-level token embeddings from playing a dominant role,
we replace candidate spans with masked tokens and use the contextual representation of the masked spans \cite{yamada-etal-2021-semantic}.
After the first step of clustering, we have coarse groups illustrated in Appendix~\ref{app:clustering}. 

\citet{michael-etal-2020-asking} suggest that we may only identify salient clusters (e.g., cardinal numbers), but cannot separate for example, different types of cardinals (e.g., number of people or number of stays). 
Thus, in the second step, we cluster examples within each cluster from the first step leveraging utterance level representation of spans (i.e., the [CLS] token of the utterance). Specifically, we identify the utterance-level representation for spans grouped from the first step. This enables us to distinguish between domains and intents as they reflect utterance-level semantics. 
For example, we may find a cluster of time information (e.g., ``11 AM'') in the first step, and the second step clustering is to differentiate between train and taxi booking time. 
Lastly, we cluster groups developed from the second step into more fine-grained types using span-level representations similar to the first step. After this multi-step clustering, we can potentially separate for instance, departure time and arrival time in train booking. This process is illustrated in Fig.~\ref{fig:multistep_illustration}. 
Each cluster represents a slot type, with slot values shown as data points.
This multi-step clustering brings an additional benefit of inducing the slot schema with hierarchy, where sub-groups in further steps belong to the same parent group.





\section{Experiments}
\label{sec:expts}
To examine the quality of our induced schema, we perform {\em intrinsic} and {\em extrinsic} evaluations. Our intrinsic evaluation compares the predicted schema with the ground truth schema by measuring their overlap in slot types and slot values. This indicates how well our induced schema aligns with the expert annotation. The extrinsic evaluation estimates the usefulness of the induced schema for downstream tasks, for which we consider dialog state tracking and response generation tasks.
Experiments are conducted on MultiWOZ \cite{eric-etal-2020-multiwoz} and SGD \cite{rastogi-etal-2020-towards} datasets following previous research. 
See Appendix~\ref{app:implementation} for implementation details. We also apply and evaluate our method for both intent and slot schema induction on realistic 
scenarios
(See Section \ref{sec:analysis}).


\paragraph{Baselines}
We compare our proposed approach with different setups against DSI \cite{min-etal-2020-dialogue}, which uses supervised tools and heuristics. We evaluate different span extraction methods including using parsers only, leveraging distance functions from LMs, and combining LMs with unsupervised PCFG.
Specifically, NP extracts all noun phrases\footnote{\url{https://spacy.io/}}, DSI cand.\ uses the same candidates phrases as DSI, and PCFG and CoreNLP \cite{manning-etal-2014-stanford} 
extract phrases from an unsupervised and supervised structure respectively by taking the smallest constituents above the leaf level. These baselines solely rely on parsers. 
For our bottom-up attention-based LM methods (Section \ref{sec:span}), we compare spans extracted using representations
from BERT \cite{devlin-etal-2019-bert}, SpanBERT \cite{joshi-etal-2020-spanbert}, TOD-BERT \cite{wu-etal-2020-tod}, and our span-based TOD pre-training from masking random spans (TOD-Span). Lastly, we combine the LMs with unsupervised PCFG structures.

Due to space constraints, we show results on MultiWOZ in this section. Observations on SGD 
can be found in the Appendix.


\subsection{Slot schema induction}
\label{sec:slot_schema_induction}
To evaluate the induced schema against ground truth, we need to 
match clusters to ground truth labels\footnote{Predicting labels for each cluster is out of the scope of this paper. 
Since there are many ways to assign labels with equal semantics to a cluster (e.g., ``food'' vs. ``restaurant type''),
we leave this to future work.}. 
Previous work on dialog schema induction either requires manual mapping from a cluster to the ground truth \cite{hudecek-etal-2021-discovering} or compares predicted slot values to its state annotation at each turn \cite{min-etal-2020-dialogue}. These can create noises and biases, hence not practical when no annotation is available. Particularly, \citet{min-etal-2020-dialogue} compare candidate spans to corresponding reference slot types at each turn, which is a small subset of the ground-truth ontology. This would overestimate the performance of schema induction since the matching is more evident and is different from defining schemas in realistic settings.
Instead, 
we simulate the process of an expert annotator mapping clusters to slot names by considering the general contextual semantics of spans in a cluster. 

\paragraph{Setup}
We consider semantic representations of ground truth clusters as labels.  
Specifically, we calculate the contextual representation of spans averaged across all spans in an induced cluster as cluster representations, and compare that with ground truth slot type representations computed in the same way. For fair comparison among different methods, we use BERT to obtain span representations.
We assign the name of the most similar slot type representation to a predicted cluster measured by cosine similarity.
If the score is lower than 0.8 \cite{min-etal-2020-dialogue}, the generated cluster is considered as noise without mapping, which simulates when a human cannot label the cluster. We report precision, recall, and F1 on the induced slot types. When the number of clusters is larger than the ground truth, multiple predicted clusters can be mapped to one slot type. This evaluation process is identical to human annotation, where the ground truth clusters serve as references (before assigning cluster labels) to predicted clusters,
but may be biased towards more clusters when more clusters are likely to cover more ground truth clusters (i.e., potentially higher recall). 
Thus we report the number of induced clusters for reference. Similarly, within each slot type, we compute the overlapping of cluster values to all ground truth slot values and report precision, recall, and F1 by fuzzy-matching scores \cite{min-etal-2020-dialogue}, averaged across all types. 

\input{tables/schema_results}

\paragraph{Results}
Table~\ref{tab:schema_results} shows the results of schema induction on slot types and slot values. All methods lead to a number of clusters within a similar range (except the slightly larger 522 clusters for DSI), indicating that the results are not biased and are comparable.
When the candidate span input to our proposed multi-step clustering is the same as the baseline DSI using POS tagging and coreference (DSI cand.), we achieve similar performance on slot type induction ($85.19$) and better results on slot values ($49.71$). 
This illustrates the effectiveness of our proposed clustering method since the only difference from the DSI baseline is clustering.
Compared to methods leveraging noun phrases (NP), or supervised parsers (CoreNLP), using an unsupervised PCFG trained on in-domain TOD data can achieve comparable or superior results.

If we extract spans using LMs only, different models perform similarly on both slot types and slot values. However, when regularized by an unsupervised PCFG structure, we observe a large performance boost especially with TOD-Span. This indicates that the unsupervised PCFG can provide complementary information to LMs. In addition, results show that further pre-training a LM at the span level is more efficient.
The better representation from span-level in-domain self-training can also be justified by a standard dialog state tracking task with few-shot or full data shown
in Appendix~\ref{app:tod_dst}. 
Detailed comparison among different LM pre-training results 
can be seen in Appendix~\ref{app:all_schema_results}. 

\subsection{Application in DST}
\input{tables/dst_results}
Now that we have mapped induced clusters to ground truth names, we can immediately evaluate DST performance by identifying slot values and types as described above.
This can be considered as a zero-shot setting. 

\paragraph{Setup}
Following \citet{min-etal-2020-dialogue}, we calculate the overlapping of the predicted slots and values with their corresponding ground truth at both the turn level and the joint level.
At each turn, a fuzzy matching score is applied on predicted values \cite{rastogi-etal-2020-towards} whose corresponding slot types are in the ground truth. On the other hand, even if a slot value is predicted correctly but its slot type does not match the ground truth, no reward is accredited. On the joint level, we calculate the score for accumulative predictions up to the current turn.


\paragraph{Results}
Table~\ref{tab:dst_results} summarizes the results for DST. Similar to the trend in schema induction, constraining an in-domain fine-tuned LM (TOD-Span) on an unsupervised structure representation (PCFG) achieves the best performance ($39.95$ on turn level), significantly outperforming a strong baseline DSI ($18.29$)\footnote{We use the provided data and model to run the DSI baseline. The reason the score is lower here than the original report is due to slot type matching (Section \ref{sec:slot_schema_induction}).}. 
We also note that because all accumulated predictions are evaluated for partial rewards instead of exact matching on all slot types in standard DST evaluation, the joint level scores are higher than the turn level from accumulative scores. See Appendix \ref{app:dsi_dst} for more detailed discussions.

\subsection{Application in response generation}
\input{tables/response_generation_results}

The above settings map latent slot clusters to ground truth analogous to expert designs so that we can evaluate the alignment with human annotations. This experiment investigates whether the induced latent schema is still useful before mapping. 

\paragraph{Setup}
We modify the model of \citet{lei-etal-2018-sequicity, zhang-etal-2020-task} by appending the predicted labels (i.e., a cluster index such as ``10-15-2'' indicating a specific slot type where each number represents a slot type from a clustering step.  This can also be considered as a hierarchical cluster label) and values to the context (e.g., ``I need a train at 7:45. [10-15-2] 7:45'' as input). The added belief state can be considered as a prior to generate responses similar to \citet{hoseini-etal-2020-simple}.
Since we do not have the mapped names of the slots, we only report the BLEU score rather than other metrics used in response generation that require entity-level matching (e.g., inform rate). 
This is a more practical setting directly evaluating on the induced schema compared to previous work \cite{min-etal-2020-dialogue}, where dialog act is modeled with 
delexicalized input utterances (\citealp{chen-etal-2019-semantically}, not feasible because ontology is required from a pre-defined schema for delexicalization).

\paragraph{Results}
Table~\ref{tab:response_generation_results} compares the performance of using no belief state (None), belief state induced by DSI, our introduced method (TOD-Span + PCFG), and ground truth. Results show that our induced schema introduces a positive inductive bias ($16.4$) compared to the baseline ($15.6$) and is close to the ground truth schema with actual slot type names. 
We conjecture that the lower performance of DSI is due to the larger number of latent types ($522$) which creates noises in model training. 
Thus, our induced slot schema is useful for downstream applications.

\section{Analysis}
\label{sec:analysis}
\paragraph{Comparison among different methods}
Our results show that in general, span-based pre-training methods outperform token-based, and continued pre-training on in-domain data is important. When regularized by PCFG structures, we observe a large performance boost on TOD-BERT and TOD-Span, however the PCFG structure does not help BERT and SpanBERT when the LM is trained on general domain data only. We speculate that the LM representation trained on generic text is not compatible with the predicted structure induced via in-domain self-supervision. This justifies our hypothesis in Section \ref{sec:span} that optimized structures from in-domain PCFG can regularize target span extraction.

In addition, we believe that the performance gap between our proposed method and previous research using rules from supervised parsers (such as NPs and coreference) is larger when the data is less biased (for example, if NP is not dominant as slot values, \citealp{du-etal-2021-qa}). Moreover, our proposed method is data-driven, indicating that the slots are determined by the dialog corpus and are more robust again label bias. If there are specific annotation requirements, we can inject inductive bias to the LM to change distribution distances \cite{shi-etal-2019-visually} or add rules to incorporate such conditions. See Appendix \ref{app:rules} for discussions.

\paragraph{Comparison among different datasets}
On MultiWOZ, our method induces 30 out of 31 slot types in the ontology except ``hospital-department'', which only appears once in the dialog corpus. For slot values, errors are mostly from low precision due to loose boundaries and semantic matching (e.g., predicting ``free wifi'', and ``include free wifi'', where the target value is ``yes''). In comparison, DSI induces 26 slot types, with similar slots mixed (such as mapping ``taxi-arriveby'' to ``taxi-leaveat''). It receives a relatively low slot value score since spans extracted using rules are not robust and compatible. 

On SGD where 82 slot types are defined in the ontology, our method induces 50 and DSI induces 72. The main reason for this low recall is similar slot types with overlapping values (such as ``media-genre'' and ``movies-genre''), and single-value slots (such as ``has-wifi'' with the value ``True'').
More importantly, SGD has a smaller utterance length, making it more difficult to map to the correct slot type without considering more context. With a magnitude more number of clusters, DSI (11992 clusters) has a higher chance to map predicted slots to target slot types which explains better performance than ours on schema induction. However, this large number of clusters make it infeasible for humans to use, and our induced schema is comparable in downstream tasks such as DST despite having a much smaller number of clusters.

On both datasets, in addition to values that can be extracted by spans, our method can also extract phrases such as ``doesn't matter'' which maps to the ``don't care'' slot value. In particular, on MultiWOZ, ``hotel-internet'' receives the lowest f1 score (0.07 with precision of 0.04 and recall of 0.35), mainly because of imprecise boundaries for low precision (e.g. ``free wifi'', ``include free wifi'', and ``offer free wifi''). It also mixes with ``free parking'' because of the context (hotel). On SGD, due to our filtering step and many slots have only one value (e.g. ``Homes-has-wifi'' and ``Homes-has-pets''), and the value (``True'') cannot be detected by spans, we received a lower schema induction score. In addition, there are 16 groups with lower matching score (< 0.8). This is particularly an issue when the number of instances is small (only 8 instances for ``home-furnished'' in total). If more instances are available, it is likely that our method can recover these missed slots due to low matching scores.

However, the schema defined here is less complicated compared to more realistic settings. For example, spans may
not be a noun phrase (such as ``until the 30th'' to distinguish from ``after the 30th'' in the utterance ``Do I have access to my premium account until the 30th or will I have to pay additional \$15 on the 29th'' to distinguish different constraints), and spans may not necessarily be meaningful arguments to intents (such as ``Can you help me to reset my password'' even though ``reset my password'' can be considered as a phrase). ABCD \cite{chen-etal-2021-action} collects more realistic TOD conversations with more in-depth discussion on finishing tasks in the shopping domain. However, they propose to leverage actions, rather than slot-value pairs as used before in slot discovery, where the actions are defined above the utterance level. 

We also apply our method on internal customer data for both intent (by applying multi-step clustering directly on utterances) and slot schema induction. Compared to MultiWOZ and SGD, schema in more realistic scenarios is more complicated and the slot boundaries are less clear. Nevertheless, our method is still effective in inducing the majority of the schema to find intents such as ``change password'' and slot types such as ``devices''. We observe similar findings on the Ubuntu dialog corpus \cite{lowe-etal-2015-ubuntu}. See more discussions in Appendix \ref{app:realistic_data}.

\input{tables/ablation_results}

\paragraph{Ablation studies}
Table~\ref{tab:ablation_results} illustrates the performance comparisons with different numbers of clustering steps, as well as input representations. Results demonstrate that compared to one-step (using masked span representation) and two-step (adding utterance representation), our three-step clustering method induces a more fine-grained schema, which is more effective for downstream tasks. 
The number of steps can be customized to real use cases depending on target granularity\footnote{More steps were also conducted but we observed lower Silhouette coefficient and lower quality in preliminary studies.}.  
In addition, if we use the original input rather than the masked phrase representation, the performance drops by a large margin ($85.71$ on slot type). This suggests that the surrounding context is more critical than the surface embeddings for schema induction, especially when the same phrase can serve different functions even in the same domain (such as locations).

\paragraph{DST Error analysis}
Suggested by the relatively high span extraction accuracy ($68.13$ F1 score) from  Appendix~\ref{app:span_extra_results}, we find that the majority of the problems in DST come from cluster mapping. This is caused by either excessive surrounding information or by the lack of context from previous turns. For instance, in the utterance ``Can I book it for 3 people'', the ``3 people'' can be mapped to either ``restaurant-book people'' or ``hotel-book people'', since we extract the contextual information from the current turn only. If more context is considered, the mapping performance including results on downstream tasks is expected to improve. Another issue is with span boundary. Even though we apply fuzzy matching, the evaluation still penalizes correct predictions (such as ``indian food'') from its ground truth (``indian''), since we do not have training signals to identify the target boundaries.

Meanwhile, we acknowledge that since we extract phrases as candidates of slot values, our DST cannot deal with other linguistic features such as coreferences and ellipses annotated in MultiWOZ and SGD. This partially explains the relatively low performance on the full zero-shot DST task. However, these features are not important for schema induction since the majority of the slot values can be found as phrases in the raw conversation, which can further be categorized into slot types. Obtaining better performance on DST is out of the scope of this paper.

\section{Conclusion}
In this paper, we propose a fully unsupervised method for slot schema induction. Compared to previous research, our method can be easily adapted to unseen domains and tasks to extract target phrases before clustering into fine-grained groups without domain constraints. We conduct extensive experiments and show that our proposed approach is flexible and effective in generating accurate and useful schemas without task-specific rules in both academic and realistic datasets. We believe that our method could also be applied to other languages (since no supervised parser, model, or heuristic is required) and tasks 
such as question answering where the answering phrase 
is not explicitly annotated \cite{min-etal-2019-discrete}. In the future, we plan to extend our method to problems with more complex structures
and data where slots are less trivial to identify.

\input{ethics}

\section*{Acknowledgements}
We thank Abhinav Rastogi from Google Research, and anonymous reviewers for their constructive suggestions. We would also thank Kenji Sagae from UC Davis for early discussions.

\bibliography{anthology,custom}
\bibliographystyle{acl_natbib}

\clearpage

\appendix

\input{appendix}

\end{document}

%% file: span_extract_alg.tex
\begin{algorithm}
\caption{Span Extraction}\label{alg:span_extract}
\begin{algorithmic}[1]
\footnotesize
\REQUIRE $\mathbf{x} = x_1, x_2, \dots, x_n$: a user utterance $\mathbf{x}$
\STATE $\mathbf{t} \gets PCFG(\mathbf{x})$ \COMMENT{A Chomsky normal form (binary) tree structure from self-supervised PCFG}
\STATE $\mathbf{a} \gets LM(\mathbf{x})$ \COMMENT{Attention distribution from a LM} 
\STATE $\mathbf{d} \gets [d_i = f(a_i, a_{i+1})$ for $i = 1, 2, \dots, n-1]$ \COMMENT{Distance between consecutive tokens using a distance function f} 
\STATE $\tau \gets$ median($\mathbf{d}$)\\
\STATE sort $\mathbf{d}$ in increasing order
\FOR {all $d_i$ in $\mathbf{d}$}
    \IF {$d_i < \tau$ and using PCFG}
        \IF {$node_i$ and $node_{i+1}$ are siblings in PCFG}
            \STATE $node_{i+1} \gets \{node_{i}, node_{i+1}$\} \COMMENT{merge nodes, assign new parents}
        \ENDIF
    \ELSIF {$d_i < \tau$}
        \STATE $w_{i+1} \gets \{w_{i}, w_{i+1}\}$ \COMMENT{merge two tokens}
    \ENDIF
\ENDFOR
\end{algorithmic}
\end{algorithm}
\vspace{-1em}

%% file: tables/schema_results.tex
\begin{table}[t]
\small
\begin{center}
\resizebox{\columnwidth}{!}{
\begin{tabular}{lccc}
\toprule
method    & \# clusters & slot type  & slot value \\
\midrule
\multicolumn{4}{l}{\textit{Baseline}}     \\
\midrule
DSI       & 522         & 87.72 & 37.18 \\
\midrule
\multicolumn{4}{l}{\textit{Parser only}}     \\
\midrule
NP        & 88          & 69.39 & 47.46 \\
DSI cand. & 113         & 85.19 & 49.71 \\
PCFG      & 339         & 91.53 & 53.62 \\
CoreNLP   & 292         & 87.72 & 54.43 \\
\midrule
\multicolumn{4}{l}{\textit{Language model only}}     \\
\midrule
BERT      & 340         & 85.71 & 55.80  \\
SpanBERT  & 343         & 89.66 & 45.21 \\
TOD-BERT  & 219         & 89.66 & 50.89 \\
TOD-Span  & 374         & 85.71 & 55.29 \\
\midrule
\multicolumn{4}{l}{\textit{Language model contrained on unsupervised PCFG}}     \\
\midrule
BERT      & 350         & 87.72 & 52.32 \\
SpanBERT  & 203         & 89.66 & 44.51 \\
TOD-BERT  & 245         & 91.53 & 48.13 \\
TOD-Span  & 290         & \textbf{96.67} & \textbf{58.71} \\
\bottomrule
\end{tabular}
}
\end{center}

\caption{\label{tab:schema_results} Schema induction results on MultiWOZ. 
TOD-Span (span-based LM further pre-trained on in-domain data) regulated by PCFG 
achieves the best performance on slot type induction and slot value induction evaluated by F1 scores. All methods (except DSI) differ only by span extraction (i.e., same clustering).
}
\vspace{-1em}
\end{table}

%% file: tables/dst_results.tex
\begin{table}[t]
\small
\begin{center}
\resizebox{\columnwidth}{!}{


\begin{tabular}{lcc}
\toprule
\multicolumn{1}{l}{method}   & turn  level         & joint level            \\
\midrule   
\multicolumn{3}{l}{\textit{Baseline}}     \\
\midrule
\multicolumn{1}{l}{DSI}      & 18.29          & 25.22              \\
\midrule
\multicolumn{3}{l}{\textit{Parser only}}     \\
\midrule
\multicolumn{1}{l}{PCFG}     & 25.43          & 32.39               \\
\midrule
\multicolumn{3}{l}{\textit{Language model only}}     \\
\midrule
\multicolumn{1}{l}{BERT}     & 24.35          & 30.18           \\
\multicolumn{1}{l}{SpanBERT} & 20.24          & 26.07          \\
\multicolumn{1}{l}{TOD-BERT} & 25.05          & 34.94         \\
\multicolumn{1}{l}{TOD-Span} & 29.72          & 38.89             \\
\midrule
\multicolumn{3}{l}{\textit{Language model contrained on unsupervised PCFG}}     \\
\midrule
\multicolumn{1}{l}{BERT}     & 23.27          & 30.09           \\
\multicolumn{1}{l}{SpanBERT} & 20.96          & 27.25                 \\
\multicolumn{1}{l}{TOD-BERT} & 27.11          & 31.92             \\
\multicolumn{1}{l}{TOD-Span} & \textbf{39.59} & \textbf{46.69}  \\

\bottomrule
\end{tabular}
 }
\end{center}

\caption{\label{tab:dst_results} DST results on MultiWOZ. We show F1 scores of turn and joint level. 
TOD-Span regularized by PCFG achieves the best performance.}
\vspace{-1em}
\end{table}

%% file: tables/response_generation_results.tex
\begin{table}[t]
\tiny
\begin{center}
\resizebox{0.6\columnwidth}{!}{
\begin{tabular}{lc}
\toprule
belief state       & BLEU \\
\midrule
None               & 15.6 \\
DSI                & 13.9 \\
TOD-Span + PCFG               & 16.4 \\
Ground truth & 17.9 \\
\bottomrule
\end{tabular}
}
\end{center}

\caption{\label{tab:response_generation_results} Response generation results on MultiWOZ. Our method introduces positive inductive bias.}
\vspace{-1em}
\end{table}

%% file: tables/ablation_results.tex
\begin{table}[t]
\small
\begin{center}
\resizebox{\columnwidth}{!}{
\begin{tabular}{lccccc}
\toprule
                  &             & \multicolumn{2}{c}{schema} & \multicolumn{2}{c}{DST} \\
method            & \# clusters & type         & value        & turn       & joint      \\
\midrule   
\multicolumn{6}{l}{\textit{Different number of clustering steps}}     \\
\midrule
one-step          & 31          & 60.87        & 39.74        & 23.58      & 30.68      \\
two-step          & 99          & 83.64        & 46.66        & 35.21      & 41.94      \\
\midrule   
\multicolumn{6}{l}{\textit{Original representation instead of masked}}     \\
\midrule
unmasked rep.      & 284         & 85.71        & 53.30         & 27.93      & 36.40       \\
\midrule   
\multicolumn{6}{l}{\textit{Three-step masked clustering}}     \\
\midrule
Three-step masked & 290         & 96.67        & 58.71        & 39.59      & 46.69      \\
\bottomrule
\end{tabular}
}
\end{center}
\vspace{-1em}
\caption{\label{tab:ablation_results} Ablation results on MultiWOZ with TOD-Span constrained on PCFG. Using masked presentation for multi-step clustering improves the performance on schema induction and DST by a large margin.}
\vspace{-1em}
\end{table}

%% file: ethics.tex
\section{Ethical Considerations}
Our intended use case is to induce the schema of raw conversations between a real user and system, where the conversation is not structured or constrained. Our experiments are done on English data, but our approach can be used for any language, especially because our method does not require any language-specific tools such as parsers which generally require a lot of labeled data. We hope that our work can reduce design and annotation cost in building dialog systems for new domains, and can inspire future research on this practical bottleneck in applications.

%% file: appendix.tex
\section{Appendices}
\label{sec:appendix}

\subsection{Implementation details}
\label{app:implementation}
For language model further pre-training, we implement our code based on \citet{wu-etal-2020-tod} where the training data and hyperparameters are kept the same. Their evaluation script is used to show results on the standard supervised dialog state tracking with the full-data and few-shot learning setting. We run all experiments on three random seeds and report the average score. The \texttt{TOD-BERT} baseline is the ``TOD-BERT-JNT-V1'' provided by \citet{wolf-etal-2020-transformers}. For span-based pre-training methods, we use the provided ``spanbert-base-cased'' model from \citet{joshi-etal-2020-spanbert} as the initial checkpoint and add a span boundary object. For random masking, we use a 15\% masking budget and sample a span length by geometric distribution with $p = 0.2$ and clip the max length to 10. For other masking methods, we follow \citet{levine-etal-2021-pmimasking} by considering n-grams of lengths 2 to 5 which appear more than 10 times in the corpus. We choose the top 10 - 20\% of n-grams by each criterion so about half of the tokens can be identified as part of correlated n-grams. We also experimented with different number of n-grams to mask and evaluate on both pre-training loss and DST results, but did not observe significant difference. We further pre-train using the same data as TOD-BERT with early stopping by prediction loss. For the attention distribution used to define our distance function, we use the eighth layer of the model suggested by \citet{kim-etal-2020-are}. We modify \citet{jin-schuler-2020-grounded} to train our unsupervised PCFG model using their suggested hyperparameters on the text input only with data cleaned by \citet{wu-etal-2020-tod}. These existing techniques, however, cannot be applied to induce schema without our proposed novel method. They only inspire us to propose an fully unsupervised method leveraging the potential benefits.
All our experiments run on eight V-100 GPUs. The training time varies from three hours to 14 hours.

For the baseline \texttt{DSI}, we run their provided public codebase on the same MultiWOZ 2.1 data and SGD dataset respectively (since each corpus has different schemas in the output space, we cannot pre-train on more task-oriented dialog data as ours), following their suggested hyperparameters on the best model DSI-GM.

For our auto-tuned multi-step clustering, we set the minimum number of samples per cluster by dividing the total number of samples by 5, 10, 15, 20, 25 and choose the best one auto-tuned by the Silhouette coefficient. A more rigorous grid search can potentially generate better performance on our tasks. All other parameters are kept as default in HDBSCAN.

For our experiments on MultiWOZ and SGD, we use the development portion of the data (following the standard separation in their original Github repositories), which represents a sample of whole corpus. MultiWOZ and SGD are commonly used task-oriented dialog datasets collected in English. On MultiWOZ, we use 7374 user utterances from the development set (1000 conversations), which covers 31 slot types. On SGD, we use 24363 user utterances (2482 conversations), which covers 82 slot types. We also report the induced schema results on the training portion of the data in Appendix \ref{app:schema_induction_training} where there are 56668 user utterances (8420 conversations) on MultiWOZ 2.1., and 164982 user utterances (16142 conversations) on SGD. We build the ground-truth ontology from the annotated corpus with slot types and values in the dialog state.  

\subsection{Algorithm}
\label{app:algorithm}
Algorithm~\ref{alg:span_extract} shows the algorithm for span extraction. For simplicity, we compare the distance from left to right for both the settings with and without PCFG structure. For using language model only, we merge tokens into phrases if their distance is small. If PCFG structure is constrained, we compare the distance between tokens and check if their corresponding nodes belong to the same parent. In practice, we implement the PCFG span extraction from bottom to top where we merge tokens into nodes from the lower level and represent the tokens with merged nodes. At each level, we compare the distance between consecutive nodes. To illustrate this process, for example in Figure~\ref{fig:pcfg_ex}, we compare the distance between the node ``modern'' and ``global cuisine'', and the distance between ``a restaurant'' and ``which'' to check if they are siblings in the same level. Since ``which'' is not merged in a lower level, itself serves as the node whereas ``a restaurant'' serves as the node for ``restaurant''. All merged phrases, with left-out unigrams, are considered as candidate extracted spans.

Algorithm~\ref{alg:clustering} shows the algorithm for auto-tuned multi-step clustering. For each step, the input to the clustering algorithm (HDBSCAN) is the embeddings of spans (or uttereances in the second step) grouped from the previous step. In other words, for each sub-groups clustered by the previous step, we further cluster the embeddings into fine-grained groups. Figure~\ref{fig:multistep_illustration} illustrates this process. The clustering algorithms returns groups of embeddings and corresponding labels (0, 1, $\dots$) and we choose the minimum number of samples per cluster based on Silhouette score. We filter clusters where the frequent spans of each sub-cluster are the same, indicating that there is only one value for this cluster. We consider the rest clusters as the input to the next step, or return as our final clusters.



\begin{algorithm*}
\caption{Auto-tuned Multi-step Clustering}\label{alg:clustering}
\begin{algorithmic}[1]
\small
\REQUIRE $\mathbf{Rep^{span}} = Rep^{span}_1, Rep^{span}_2, \dots, Rep^{span}_n$: masked span representation (hidden states of LM by replacing extracted spans with [MASK] token)
\REQUIRE $\mathbf{Rep^{utt}} = Rep^{utt}_1, Rep^{utt}_2, \dots, Rep^{utt}_n$: utterance-level representation (hidden states of LM on [CLS] token)
\REQUIRE $\mathbf{min\_nums}$: a list of candidate values to set for minimum samples for cluster. This is not sensitive to the clustering results.
\STATE $input\_embeddings \gets Rep^{span}$
\STATE $clusters \gets {input\_embeddings}$
\FOR {$step_i$ in multi-steps}
    \FOR {$input\_embeddings_i$ in $clusters$}
        \STATE $clusters_i \gets max\_i\{silhouette\_score(HDBSCAN(input\_embeddings_i, min\_num_i))\}$ \COMMENT{Clustered group of embeddings}
        \IF {$step\_i = 1$}
            \IF {all sub-clusters share the same frequent span}
                \STATE ignore $input\_embeddings_i$, continue the for loop \COMMENT{filter clusters with only one value}
            \ENDIF
            \STATE $clusters_i \gets$ corresponding $Rep^{utt}$ for each item in $clusters_i$ \COMMENT{Use utterance-level representation for the second step clustering}
        \ENDIF
    \ENDFOR
    \STATE $clusters \gets$ \{$clusters_i$ for all i in the current step\}
\ENDFOR

\end{algorithmic}
\end{algorithm*}


\subsection{Supervised DST results}
\label{app:tod_dst}
\input{tables/tod_lm_results}

\input{tables/tod_lm_results_few}
\citet{wu-xiong-2020-probing} suggest that further pre-training on TOD data \cite{wu-etal-2020-tod} helps generating better utterance-level representation, but less so for other features such as slots. To encourage better span-level representation, we further pre-trained a SpanBERT model on TOD data by masking spans based on frequency, Pointwise Mutual Information (PMI), symmetric conditional probability (SCP, \citealp{downey-etal-2007-locating}), and segmented PMI \cite{levine-etal-2021-pmimasking} following recent research, together with randomly masking contiguous random spans. Implementation details can be found in Appendix \ref{app:implementation}. Here we evaluate different pre-trained methods on the standard DST benchmark. 

Table~\ref{tab:tod_dst} and Table~\ref{tab:tod_dst_few} shows the performance of supervised DST performance evaluated on joint accuracy and slot accuracy with the full data and few-shot data (1 - 10\%), respectively. Note that this was not used to choose the best model to perform schema induction and related tasks. These results compare different pre-training methods to illustrate the quality of the initial checkpoints on a more standard benchmark. As shown similarly in recent work, TOD-BERT can only show marginal improvement over BERT averaged over different random seeds. Meanwhile, \texttt{SpanBERT} when used as an initial checkpoint is not stable at downstream DST tasks even if multiple random seeds were tested. However, after further pre-training on task-oriented dialog dataset, \texttt{TOD-Span} achieve significantly better performance in both the few-shot and full-data setting. When comparing different span masking methods, random masking (\texttt{TOD-Span}) is quite effective. Although \texttt{freq} and \texttt{PMI\_seg} achieves better performance (over the naive \texttt{PMI}), the improvement is not large. We conjecture that this might be due to that compared to general domains and tasks with more diverse prediction space such as question answering, the number of task-relevant phrases in task-oriented dialog is limited.  

\input{tables/span_extrac_results}

\subsection{Span Extraction Results}
\label{app:span_extra_results}

\begin{figure*}[ht]
\centering
\includegraphics[width=\linewidth]{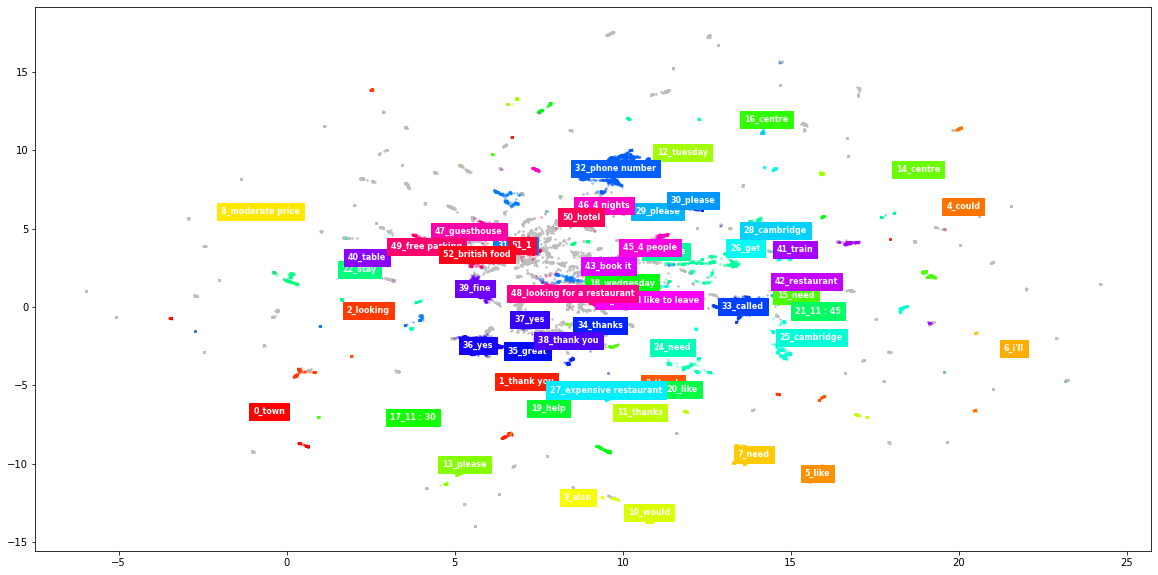}
\caption{Clustering after first step. Grey labels are outliers detected by HDBSCAN. The numbers in each group represent a latent cluster label, and the texts represent the most frequent phrase in cluster.}
\label{fig:clustering_first_step}
\end{figure*}

Table~\ref{tab:span_extrac_results} shows the recall for span extraction results. We manually annotate 200 user utterances so that acceptable span boundaries would not be penalized. For instance, given the utterance ``I need to book a hotel in the east that has 4 stars'', instead of the DST annotation ``hotel-starts: 4'' and ``hotel-area: east'' together with coreference and annotation errors that cannot be detected from the context, we manually annotate the candidate spans as [``in the east'', ''the east'', ''east''] and [''4 stars'', ''has 4 stars'', ''4''] which relaxes the rigid requirement of strict matching of slot values. Compared to fuzzy matching, this evaluation is cleaner. Because of the annotation errors and coreference that a value does not appear in the current utterance, the ground truth score is $78.83$. Similar to our schema induction and DST evaluation results, we observe that constraining on predicted structures can increase model performance. In particular, using an in-domain self-supervised PCFG structure and achieve similar or even better performance than using a supervised parser. 
We only evaluate recall here because there are non-meaningful spans extracted, and is not important to downstream tasks since they are potentially filtered by our clustering method.

\subsection{Clustering}
\label{app:clustering}
Figure~\ref{fig:clustering_first_step} shows the clustering results after the first step.  This shows that we can get some coarse clusters with non-meaningful groups (such as ``thank you''). Some slot types (such as day of the week as ``wednesday'') are not distinguished by their domain and intent. Further clustering can generate more fine-grained schema.

In addition, from empirical analysis, we found that meaningless spans extracted together with meaningful ones from the previous stage may add noises in the process. To study its influence by filtering out noisy clusters, we automatically examine clusters and their corresponding sub-clusters from the first two steps based on the assumption that valid slot types include more than one slot value. We choose one here because if one cluster is dominated by examples such as ``thank you'' with a few other instances such as ``thanks'', the latter can be considered as outliers from our clustering method. Afterwards there is only one value left. We can also choose to filter out clusters with more than one slot value, which may result in lower recall.
Since the goal of schema induction is to build a complete ontology with high recall, noisy groups are actually acceptable. In other words, we observed similar performance before and after filtering out such noisy cluster since the cluster mapping step would assign a low score to such groups from cluster embedding representations (Section \ref{sec:slot_schema_induction}), which is similar to how human experts would ignore a cluster of meaningless spans.

\subsection{Schema induction on training portion}
\label{app:schema_induction_training}
\input{tables/induce_training}
Since our goal is to induce the schema of a corpus without using any labeled data, there is no major difference in whether the schema is induced on the training set of MultiWOZ or the development set. The main difference is the number of utterance where the training data is ten times larger than the development data. Here we show the results for reference. Table~\ref{tab:induce_training} demonstrates that despite our much smaller number of clusters, our method achieves significantly better performance than the \texttt{DSI} baseline on both schema induction and DST.

\subsection{SGD results}
\label{app:sgd_results}
\input{tables/SGD_results}
Table~\ref{tab:SGD_results} shows the results for schema induction and DST on the SGD dataset. We conjecture that the similar performance results with the strong \texttt{DSI} baseline is due to large difference in cluster numbers. Intuitively, with a larger number of clusters, each group with fewer examples can be mapped to the ground truth embeddings correctly. On the other hand, if different slot types are mixed into one cluster, all slot values are assigned an inaccurate name.
Another potential reason is that compared to MultiWOZ, SGD dataset requires more contextual information (SGD has less average tokens per turn and more turns per dialogue). Thus the mapping from relatively noisy clusters 
to ground truth creates errors for downstream tasks, especially that the evaluation metric require exact match of slot types.

However, the results are still comparable. Although predicting a magnitude smaller number of clusters to be less favoured in evaluation, our method still achieves similar or better performance.

\subsection{Comparison to DSI on DST}
\label{app:dsi_dst}
We note that the DST results on MultiWOZ for DSI is lower than that reported in \citet{min-etal-2020-dialogue}. As shown in Section \ref{sec:slot_schema_induction}, the original number was reported by mapping predicted slot types to target ontology at the turn level (before accumulating for the final prediction), where a small subset is used. This process makes mapping more evident (for example, instead of mapping a predicted slot type to the target 30 slot types, it only compares a slot type to one of two slot types that appear in the reference). Hence, it overestimates the performance and is very different from how a human expert would assign labels when inducing the schema for a new corpus without (turn-level) annotation. Since DST is evaluated to make sure that the slot type matches, an incorrect slot type matching would result in a 0 true positive score. The actual performance in our experiments is thus lower. 

In addition, we follow the exactly same settings including training and evaluation scripts (on DST) with their provided pre-processed span-level data and suggested hyperparameters. We use the same metrics and scripts to evaluate all methods. Accordingly, all the numbers reported in Table \ref{tab:dst_results} are fair and comparable.

Lastly, since we use fuzzy matching scores \cite{rastogi-etal-2020-towards, min-etal-2020-dialogue}, turn-level performance is accumulated to the joint level. For that reason, different from joint goal accuracy commonly used where all slot types and values are required to be exactly match, partial true positives are counted again in future turns. For example, if the current turn predicts ``train leave-at: 10'' with the target dialog state ``train leave-at: 10:00'', even if the next turn predicts nothing correctly, this partial score is counted in the joint level score in the next turn. This procedure follows the setting of \citet{min-etal-2020-dialogue}. In fact, in their reported performance of DSI-GM on MultiWOZ 2.1 with precision and recall of 52.5, 39.3, and 49.2, 43.2 at turn and joint level respectively, the actual F1 scores are actually 45.0 for the turn level and 46.0 at the joint level. Similar to ours, they also received higher score on the joint level due to accumulative partial scores (by calculating F1 using their reported precision and recall scores directly). Since we follow the same evaluation script and metrics, the results and conclusion we have in our experiments comparing different methods are comparable. 

\subsection{Further analysis of schema induction among datasets (including realistic data)}
\label{app:realistic_data}

When we apply our method on internal customer data for slot schema induction, we follow the same pipeline introduced in Section \ref{sec:method}. For intent schema induction, we consider both the system turn and utterance turn as the context to our multi-step clustering to find schema with the hierarchy. Because our method is data-driven and does not require heuristics, it can induce expected slots explained before (e.g. ``until the 30th''). We observed empirically satisfying performance but the results cannot be reported publicly because of restrictions. Therefore, we only report results on the public datasets to compare to previous research, as well as to inspire follow-up works for comparison. 

We also applied our approach to the Ubuntu dialog corpus \cite{lowe-etal-2015-ubuntu}. Compared to general TOD systems where a user and an knowledgeable agent communicate with each other, this data is collected from online forms to discuss technical issues. The utterances are less conversational, and include coding scripts, making it very noisy. We experiment on this more realistic dataset only for reference, since it is significantly different from building a TOD systems to interact with real users where schema is critical. We sample 8k utterances from the training data, and apply our method on both the intent level and the slot level. On the intent level, our method generates 70 clusters from the first step, and 154 clusters after three steps. Apart from greetings (which appear very frequently), we can induce intents such as suggesting one question is off topic (e.g. ``this is a support channel; please leave and go to xxx channel''). There are also some more evident intent clusters such as suggested command line, suggested url, and questions for installations in a specific setup (e.g. ``how to install firefox on 64 bit''). When the input sentence is long with mixed intents, our method may group these into one large cluster (such as providing suggestions to a specific problem, which is more similar to dialog act). We may choose to mix slot- and utterance-level clustering to solve such an issue by treating each complete segment in an utterance as a long span. On the slot level induction, our method generates 36 clusters from the first step, and 287 clusters after three steps. Our method can induce slot types such as ``ubuntu verision'' and ``software name''. However, compared to MultiWOZ and SGD, the induced slots are much nosier with lower precision where meaningless verbs (e.g. ``set up'' are grouped). Meanwhile, there a many other slot types that are not meaning such as a cluster regarding part of a path (e.g., ``/var/''), which may be due to that we use the same LM trained on TOD dat which does not handle code scripts. Further in-domain pre-training within the Ubuntu dialog corpus may solve this issue. To conclude, even though this dataset is noisy and different from TOD, our method is still applicable to discover useful schema on both the intent and the slot level without any supervision.

 
\subsection{Applying induced schema on testing data}
After inducing the schema on the training data, we may apply the induced schema directly to a different set of data (such as testing data) for downstream applications (such as DST). Since we already induced slot clusters and mapped them to ground truth, we do not need to follow the same span extraction before clustering again. Alternatively, 
we adopt the following procedure.
We extract all candidate phrases in the same way, but instead of clustering, we map the extracted phrases to clustered groups. Specifically, similar to mapping induced latent clusters to ground truth groups in schema induction, we find the most similar latent cluster to the candidate in the contextualized embedding space, and assign the cluster name to the phrase as its slot type. We observed that even though the schema is not induced on the testing data, the performance on both turn and joint level maintains (36.58 and 48.98).

\subsection{Related work in schema induction of other natural language processing tasks}
\label{app:related_work}
Similar to grammar induction and unsupervised parsing, schema induction can help to eliminate the time-consuming manual process and serves as the first step to build a large corpus \cite{klein-manning-2002-generative, klasinas-etal-2014-semeval}. Related tasks include event type induction \cite{huang-etal-2016-liberal, huang-etal-2018-zero}, semantic frame induction \cite{yamada-etal-2021-semantic}, and semantic role induction \cite{lang-lapata-2010-unsupervised, michael-zettlemoyer-2021-inducing}. Relationship in these tasks such as predicate and head or patient and agent are relatively evident compared to that in conversational dialog. In addition, most of previous research requires either strong statistical assumptions based on pre-defined parsers, or existing ontologies and annotations for some seen types, and formulate the problem similar to word sense disambiguation on predicate-object pairs \cite{shen-etal-2021-corpus}. In contrast, our method does not require any formal syntactic or semantic supervision.  

\subsection{Incorporating task-specific annotation requirements for schema induction}
\label{app:rules}
Our method is data-driven, indicating that if two tokens appear frequently (thus form a span), it might be a good idea to consider them as a slot together. Our motivation here is to induce the most probabilistic schema based on distributed representations. Incorporating annotation requirement is not specific to schema induction from corpus, and is a broader concept of neuro-symbolic integration by merging symbolic rules with connectionist models like neural networks. 

However, if there is a specific requirement, we can either inject inductive bias similar to \citet{shi-etal-2019-visually, kim-etal-2020-are} to change the attention distribution (so that the requirement-specific bias can result in smaller or larger divergence explicitly). We can also add such requirements as rules directly on certain spans. In this way, we can incorporate the requirements. In comparison, previous methods relying on supervised parser are not applicable.

\subsection{Detailed schema induction results}
\label{app:all_schema_results}
\input{tables/all_schema_results}
Table~\ref{tab:all_schema_results} shows detailed results comparison on different proposed methods on schema induction. All methods result in a similar number of clusters, while span-based further pre-training methods constrained on unsupervised PCFG structures achieve the best performance overall.

%% file: tables/tod_lm_results.tex
\begin{table}[t]
\small
\begin{center}
\resizebox{\columnwidth}{!}{
\begin{tabular}{lcc}
\toprule
Model & Joint Acc.   & Slot Acc.  \\ 
\midrule
BERT & 45.6 & 96.6 \\
SpanBERT & 1.5 & 81.1 \\
ToD-BERT & 46.0 & 96.6 \\
\midrule
\multicolumn{3}{l}{\textit{Span-based model trained on TOD data}}     \\
\midrule
TOD-Span & 49.0 & 96.9 \\
freq & \textbf{49.7} & \textbf{97.0} \\
freq w/o stop & 47.3 & 96.8 \\
PMI & 48.7 & 96.9 \\
PMI\_seg & 49.4 & \textbf{97.0} \\
SCP & 48.3 & 96.8 \\

\bottomrule
\end{tabular}
}
\end{center}
\caption{\label{tab:tod_dst} Supervised DST results with the full-data setting. Results show that span-based methods outperform token-based pre-training methods, and this improvement is not from the initial checkpoint. Different masking methods achieve similar performance.}
\end{table}

%% file: tables/tod_lm_results_few.tex
\begin{table}[t]
\small
\begin{center}
\resizebox{\columnwidth}{!}{
\begin{tabular}{llcc}
\toprule
data & Model & Joint Acc.   & Slot Acc.  \\ 
\midrule
\multirow{4}{*}{1\%} & BERT     & 6.4           &  84.4         \\
                     & SpanBERT &   3.6         &    82.6       \\
                     & TOD-BERT &   7.9         &    84.9       \\
                     & TOD-Span &   9.9        &   86.0       \\
\midrule
\multirow{4}{*}{5\%} & BERT   &   19.6         &  92.0         \\
                     & SpanBERT &   5.6         &   83.9        \\
                     & TOD-BERT &   20.9         &  91.0         \\
                     & TOD-Span &   28.2         &  93.9         \\
\midrule                     
\multirow{4}{*}{10\%} & BERT &  32.9          &   94.7        \\
                     & SpanBERT &   11.8         &    85.6       \\
                     & TOD-BERT &   30.2         &    93.5       \\
                     & TOD-Span &   38.6         &    95.5       \\

\bottomrule
\end{tabular}
}
\end{center}
\caption{\label{tab:tod_dst_few} Supervised DST results with few-shot training data. Similar to the full-data setting, span-based methods achieve significantly better performance than token-based further pre-training methods.}
\end{table}

%% file: tables/span_extrac_results.tex
\begin{table}[t]
\small
\begin{center}
\resizebox{\columnwidth}{!}{
\begin{tabular}{lccc}
\toprule
Model & R (LM only)   & R (+ supervised) & R (+ unsupervised)  \\ 
\midrule
NP & 62.13 &  &\\
BERT & 62.30 & 62.05 & 64.30 \\
SpanBERT & 58.43 & 64.60 & 62.52 \\
TOD-BERT & 54.15 & 60.88 & 65.05 \\
TOD-Span & 64.21 & 67.22 & \textbf{68.13} \\
\midrule
Ground Truth & 78.83 & &\\

\bottomrule
\end{tabular}
}
\end{center}
\caption{\label{tab:span_extrac_results} Span extraction results on manually labeled utterances. Results show that constrained on unsupervised PCFG structure, our span-based further pre-training method TOD-Span achieves the best recall ($68.13$), close to the ground truth performance ($78.83$)}
\end{table}

%% file: tables/induce_training.tex
\begin{table}[ht]
\small
\begin{center}
\resizebox{\columnwidth}{!}{
\begin{tabular}{lccccc}
\toprule
                  &             & \multicolumn{2}{c}{schema} & \multicolumn{2}{c}{DST} \\
method            & \# clusters & type         & value        & turn       & joint      \\
\midrule   

DSI          & 4981          & 95.08      & 43.23      & 21.10      & 28.14     \\
Ours          & 374          & 93.33        & \textbf{47.32}        & \textbf{37.64}      & \textbf{44.74}      \\

\bottomrule
\end{tabular}
}
\end{center}

\caption{\label{tab:induce_training} Results for schema induction and DST when the schema is induced on the training portion of MultiWOZ data. Our method significantly outperforms the strong DSI baseline.}
\end{table}

%% file: tables/SGD_results.tex
\begin{table}[ht]
\small
\begin{center}
\resizebox{\columnwidth}{!}{
\begin{tabular}{lccccc}
\toprule
                  &             & \multicolumn{2}{c}{schema} & \multicolumn{2}{c}{DST} \\
method            & \# clusters & type         & value        & turn       & joint      \\
\midrule   

DSI          & 11992          & 92.21      & 46.19       & 27.23      & 26.24     \\
Ours          & 806          & 77.04        & 47.50        & 26.01      & 26.50      \\

\bottomrule
\end{tabular}
}
\end{center}

\caption{\label{tab:SGD_results} Schema induction and DST results on SGD dataset. Results suggests that our method achieves comparable or better performance than the strong DSI baseline even though our number of clusters is a magnitude smaller. See text for analysis. }
\end{table}

%% file: tables/all_schema_results.tex
\begin{table*}[ht]
\small
\begin{center}
\resizebox{\linewidth}{!}{
\begin{tabular}{lccccccc}
\cline{3-8}
                              \multicolumn{2}{l}{}            & \multicolumn{3}{c}{slot type}                                                         & \multicolumn{3}{c}{slot value}                                                        \\ 
\midrule
method & \# clusters & precision & recall & f1 & precision & recall & f1 \\
\toprule
\multicolumn{4}{l}{\textit{Baseline}}     \\
\midrule
DSI                          & 522                              & 96.15                          & 80.65                       & 87.72                   & 41.53                          & 57.40                       & 37.18                   \\
\midrule
\multicolumn{4}{l}{\textit{Parser only}}     \\
\midrule
NP                           & 88                               & 94.44                          & 54.84                       & 69.39                   & 42.26                          & 67.80                       & 47.46                   \\
DSI cand.                    & 113                              & \textbf{100.00}                & 74.19                       & 85.19                   & 56.46                          & 60.80                       & 49.71                   \\
PCFG                         & 339                              & 96.43                          & 87.10                       & 91.53                   & 62.14                          & 58.01                       & 53.62                   \\
CoreNLP                      & 292                              & 96.15                          & 80.65                       & 87.72                   & 57.80                          & 63.18                       & 54.43                   \\
\midrule
\multicolumn{4}{l}{\textit{Language model only}}     \\
\midrule
BERT                         & 340                              & 96.00                          & 77.42                       & 85.71                   & 62.11                          & 58.60                       & 55.80                   \\
SpanBERT                     & 343                              & 96.30                          & 83.87                       & 89.66                   & 56.34                          & 51.95                       & 45.21                   \\
TOD-BERT                     & 219                              & 96.30                          & 83.87                       & 89.66                   & \textbf{63.58}                          & 57.64                       & 50.89                   \\
TOD-Span                     & 374                              & 96.00                          & 77.42                       & 85.71                   & 54.88                          & 69.13                       & 55.29                   \\
freq                         & 100                              & 93.33                          & 45.16                       & 60.87                   & 47.31                          & 63.32                       & 45.97                   \\
freq w/o stop                & 337                              & 95.65                          & 70.97                       & 81.48                   & 48.63                          & 63.66                       & 48.27                   \\
PMI                         & 369                              & \textbf{100.00}                & 80.65                       & 89.29                   & 53.97                          & \textbf{73.60}              & 56.38                   \\
PMI\_seg                     & 551                              & 96.55                          & 90.32                       & 93.33                   & 60.37                          & 66.68                       & 58.33                   \\
SCP                          & 374                              & 96.00                          & 77.42                       & 85.71                   & 55.06                          & 61.23                       & 51.78                   \\
\midrule
\multicolumn{4}{l}{\textit{Language model contrained on unsupervised PCFG}}     \\
\midrule
BERT                         & 350                              & 96.15                          & 80.66                       & 87.72                   & 58.85                          & 57.49                       & 52.32                   \\
SpanBERT                     & 203                              & 96.30                          & 83.87                       & 89.66                   & 60.54                          & 48.23                       & 44.51                   \\
TOD-BERT                     & 245                              & 96.43                          & 87.10                       & 91.53                   & 55.40                          & 57.26                       & 48.13                   \\
TOD-Span                     & 290                              & \textbf{100.00}                & \textbf{93.55}              & \textbf{96.67}          & 61.34                 & 67.26                       & \textbf{58.71}          \\
freq                         & 379                              & \textbf{100.00}                & 83.87                       & 91.23                   & 56.67                          & 68.19                       & 57.19                   \\
freq w/o stop                & 315                              & 96.55                          & 90.32                       & 93.33                   & 56.40                          & 66.43                       & 53.74                   \\
PMI                          & 335                              & 96.55                          & 90.32                       & 93.33                   & 57.90                          & 67.50                       & 56.91                   \\
PMI\_seg                     & 275                              & 96.55                          & 90.32                       & 93.33                   & 55.19                          & 65.04                       & 54.54                   \\
SCP                          & 290                              & \textbf{100.00}                & 90.32                       & 94.92                   & 53.62                          & 65.31                       & 53.00         \\

\bottomrule
\end{tabular}
}
\end{center}

\caption{\label{tab:all_schema_results} Schema induction results for different proposed methods. High precision scores indicate that the clusters are relatively clean without noises that result in fuzzy representations for mapping. }
\end{table*}